\documentclass[letterpaper]{article} % DO NOT CHANGE THIS
\usepackage{aaai23}  % DO NOT CHANGE THIS
\usepackage{times}  % DO NOT CHANGE THIS
\usepackage{helvet}  % DO NOT CHANGE THIS
\usepackage{courier}  % DO NOT CHANGE THIS
\usepackage[hyphens]{url}  % DO NOT CHANGE THIS
\usepackage{graphicx} % DO NOT CHANGE THIS
\usepackage{natbib}  % DO NOT CHANGE THIS AND DO NOT ADD ANY OPTIONS TO IT
\usepackage{caption} % DO NOT CHANGE THIS AND DO NOT ADD ANY OPTIONS TO IT
\usepackage{algorithm}
\usepackage{algorithmic}

\usepackage{multirow}
\usepackage{subfigure}
\usepackage{algorithm}
\usepackage{xspace}
\usepackage{dsfont}
\usepackage{epsfig}
\usepackage{epstopdf}
\usepackage{booktabs}
\usepackage{indentfirst}
\usepackage{amsmath}
\usepackage{amsthm}
\usepackage{amsfonts}
\usepackage{bm}
\usepackage{listings}
\usepackage{tabu}
\usepackage{longtable}
\usepackage{threeparttable}
\usepackage{marvosym}
\usepackage{tabularx}
\usepackage{cite}
\usepackage{color}

\newtheorem{mydef}{Definition}
\newcommand{\name}{PDFormer\xspace}
\newcommand{\paratitle}[1]{\vspace{1.5ex}\noindent\textbf{#1}}
\newcommand{\ie}{\emph{i.e.,}\xspace}

\newcommand{\eg}{\emph{e.g.,}\xspace}

\newcommand{\ignore}[1]{}

\newcommand{\pfour}{PeMS04\xspace}
\newcommand{\pseven}{PeMS07\xspace}
\newcommand{\peight}{PeMS08\xspace}
\newcommand{\taxi}{NYCTaxi\xspace}
\newcommand{\bike}{CHIBike\xspace}
\newcommand{\bj}{T-Drive\xspace}

\newcommand{\tah}{$\mathrm{TAH}$\xspace}
\newcommand{\ssa}{$\mathrm{SSA}$\xspace}
\newcommand{\tsa}{$\mathrm{TSA}$\xspace}
\newcommand{\gsah}{$\mathrm{GeoSAH}$\xspace}
\newcommand{\ssah}{$\mathrm{SemSAH}$\xspace}
\newcommand{\gssa}{$\mathrm{GeoSSA}$\xspace}
\newcommand{\sssa}{$\mathrm{SemSSA}$\xspace}
\newcommand{\dft}{$\mathrm{DFT}$\xspace}

\usepackage{newfloat}
\usepackage{listings}
\DeclareCaptionStyle{ruled}{labelfont=normalfont,labelsep=colon,strut=off} % DO NOT CHANGE THIS
\floatstyle{ruled}
\newfloat{listing}{tb}{lst}{}
\floatname{listing}{Listing}
\title{\name: Propagation Delay-Aware Dynamic Long-Range Transformer for Traffic Flow Prediction}
\author{
    Jiawei Jiang,\textsuperscript{\rm 1}\equalcontrib\,
    Chengkai Han,\textsuperscript{\rm 1}\equalcontrib\,
    Wayne Xin Zhao,\textsuperscript{\rm 4}\,
    Jingyuan Wang\textsuperscript{\rm 1,\rm 2,\rm 3}\thanks{Corresponding author: jywang@buaa.edu.cn\\}
}
\affiliations{
    \textsuperscript{\rm 1}School of Computer Science and Engineering, Beihang University, Beijing, China\\
    \textsuperscript{\rm 2}Pengcheng Laboratory, Shenzhen, China\\
    \textsuperscript{\rm 3}School of Economics and Management, Beihang University, Beijing, China\\
    \textsuperscript{\rm 4}Gaoling School of Artificial Intelligence, Renmin University of China, Beijing, China\\
    \{jwjiang, ckhan, jywang\}@buaa.edu.cn; batmanfly@gmail.com
}

\begin{document}

\maketitle

\begin{abstract}
As a core technology of Intelligent Transportation System, traffic flow prediction has a wide range of applications. The fundamental challenge in traffic flow prediction is to effectively model the complex spatial-temporal dependencies in traffic data. Spatial-temporal Graph Neural Network (GNN) models have emerged as one of the most promising methods to solve this problem. However, GNN-based models have three major limitations for traffic prediction: i) Most methods model spatial dependencies in a static manner, which limits the ability to learn dynamic urban traffic patterns; ii) Most methods only consider short-range spatial information and are unable to capture long-range spatial dependencies; iii) These methods ignore the fact that the propagation of traffic conditions between locations has a time delay in traffic systems. To this end, we propose a novel \underline{P}ropagation \underline{D}elay-aware dynamic long-range trans\underline{Former}, namely \name, for accurate traffic flow prediction. Specifically, we design a spatial self-attention module to capture the dynamic spatial dependencies. Then, two graph masking matrices are introduced to highlight spatial dependencies from short- and long-range views. Moreover, a traffic delay-aware feature transformation module is proposed to empower \name with the capability of explicitly modeling the time delay of spatial information propagation. Extensive experimental results on six real-world public traffic datasets show that our method can not only achieve state-of-the-art performance but also exhibit competitive computational efficiency. Moreover, we visualize the learned spatial-temporal attention map to make our model highly interpretable.
\end{abstract}
% Specifically, we design a spatial self-attention module to capture the dynamic spatial dependencies. Then, two graph masking matrices are introduced to capture the short-range and long-range spatial dependencies simultaneously.

% to highlight spatial dependencies from short- and long-range views.

% \keywords{Traffic Prediction, Self-Attention, Graph Transformer}

\section{Introduction}

In recent years, rapid urbanization has posed great challenges to modern urban traffic management. As an indispensable part of modern smart cities, intelligent transportation systems (ITS)~\cite{intro} have been developed to analyze, manage, and improve traffic conditions (\eg reducing traffic congestion). As a core technology of ITS, \emph{traffic flow prediction}~\cite{intro2} has been widely studied, aiming to predict the future flow of traffic systems based on historical observations. It has been shown that accurate traffic flow prediction can be useful for various traffic-related applications~\cite{libcity}, including route planning, vehicle dispatching, and congestion relief.

For traffic flow prediction, the fundamental challenge is to effectively capture and model the complex and dynamic spatial-temporal dependencies of traffic data~\cite{intro3}. Many attempts have been made in the literature to develop various deep learning models for this task. As early solutions, convolutional neural networks (CNNs) were applied to grid-based traffic data to capture spatial dependencies, and recurrent neural networks (RNNs) were used to learn temporal dynamics~\cite{STResNet, dmvstnet}. Later, graph neural networks (GNNs) were shown to be more suited to model the underlying graph structure of traffic data~\cite{DCRNN, STGCN}, and thus GNN-based methods have been widely explored in traffic prediction~\cite{GWNET,STSGCN,MTGNN,STFGNN,STGODE,STG-NCDE}. %Recently, the attention mechanism have been introduced in traffic prediction due to its ability to capture the dynamic spatial-temporal dependencies~\cite{GMAN, astgnn}.  

%The main challenge in traffic prediction is modeling the data's complex and dynamic spatial-temporal dependencies. Many attempts have been made in the literature to develop deep learning models to solve this problem. Early on, convolutional neural networks (CNNs) were applied to grid-based traffic data to capture spatial dependencies, and recurrent neural networks (RNNs) were used to learn temporal dynamics~\cite{STResNet, dmvstnet}. Later, researchers found that graph neural networks (GNNs) were suitable for modeling the underlying graph structure of traffic data. Therefore, GNN-based methods have been widely explored in traffic prediction~\cite{DCRNN,STGCN,GWNET,STSGCN,MTGNN,STFGNN,STGODE,STG-NCDE}. %Recently, the attention mechanism have been introduced in traffic prediction due to its ability to capture the dynamic spatial-temporal dependencies~\cite{GMAN, astgnn}.  

\begin{figure}[t]
    \centering
    \includegraphics[width=1\columnwidth]{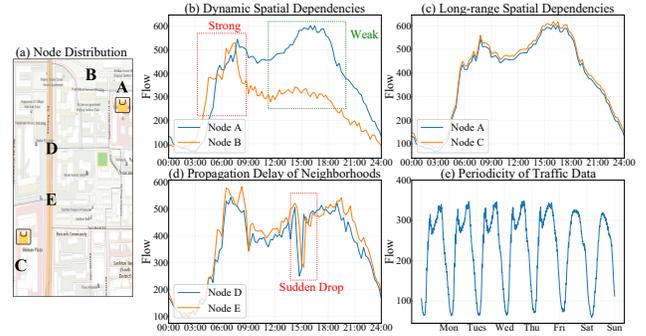}
    \caption{The Findings about Traffic Prediction.}
    \label{fig:intro}
\end{figure}

% \begin{figure}[t]
%     \centering
%     \includegraphics[width=0.95\columnwidth]{figure/intro_full.png}
%     \caption{The Findings about Traffic Prediction.}
%     \label{fig:intro}
% \end{figure}

% \begin{figure}[t]
%     \centering
%     \includegraphics[width=0.9\columnwidth]{figure/intro2.pdf}
%     \caption{The Findings about Traffic Prediction.}
%     \label{fig:intro}
% \end{figure}

Despite the effectiveness, GNN-based models still have three major limitations for traffic prediction. Firstly, the spatial dependencies between locations in a traffic system are highly \emph{dynamic} instead of being static, which are time-varying as they are affected by travel patterns and unexpected events. For example, as shown in Fig.~\ref{fig:intro}(b), the correlation between nodes $A$ and $B$ becomes stronger during the morning peak and weaker during other periods. While, existing methods model spatial dependencies mainly in a static manner (either predefined or self-learned), which limits the ability to learn dynamic urban traffic patterns. Secondly, due to the functional division of the city, two distant locations, such as nodes $A$ and $C$ in Fig.~\ref{fig:intro}(c), may reflect similar traffic patterns, implying that the spatial dependencies between locations are \emph{long-range}. Existing methods are often designed locally and unable to capture long-range dependencies. For example, GNN-based models suffer from over-smoothing, making it difficult to capture long-range spatial dependencies. Thirdly, the effect of \emph{time delay} might occur in the spatial information propagation between locations in a traffic system. For example, when a traffic accident occurs in one location, it will take several minutes (a delay) to affect the traffic condition in neighboring locations, such as nodes $D$ and $E$ in Fig.~\ref{fig:intro}(d). However, such a feature has been ignored in the immediate message passing mechanism of typical GNN-based models~\cite{wang2017community}.

To address the above issues, in this paper, we propose a \underline{P}ropagation \underline{D}elay-aware dynamic long-range trans\underline{Former} model, namely \name, for traffic flow prediction. As the core technical contribution, we design a novel spatial self-attention module to capture the dynamic spatial dependencies. This module incorporates local geographic neighborhood and global semantic neighborhood information into the self-attention interaction via different graph masking methods, which can simultaneously capture the short- and long-range spatial dependencies in traffic data. Based on this module, we further design a delay-aware feature transformation module to integrate historical traffic patterns into spatial self-attention and explicitly model the time delay of spatial information propagation. Finally, we adopt the temporal self-attention module to identify the dynamic temporal patterns in traffic data. In summary, the main contributions of this paper are summarized as follows:

\begin{itemize}
\item We propose the \name model based on the spatial-temporal self-attention mechanism for accurate traffic flow prediction. Our approach fully addresses the issues caused by the complex characteristics from traffic data, namely dynamic, long-range, and time-delay. 
\item We design a spatial self-attention module that models both local geographic neighborhood and global semantic neighborhood via different graph masking methods and further design a traffic delay-aware feature transformation module that can explicitly model the time delay in spatial information propagation.
\item We conduct both multi-step and single-step traffic flow prediction experiments on six real-world public datasets. The results show that our model significantly outperforms the state-of-the-art models and exhibits competitive computational efficiency. Moreover, the visualization experiments show that our approach is highly interpretable via the learned spatial-temporal attention.
\end{itemize}

\section{PRELIMINARIES}
In this section, we introduce some notations and formalize the traffic flow prediction problem.
\subsection{Notations and Definitions}
% In this paper, we both consider highway traffic flow (graph-based) and citywide traffic flow (grid-based) data. Therefore, we make the following definitions:

\begin{mydef}[Road Network]
We represent the \emph{Road Network} as a graph $\mathcal{G} = (\mathcal{V}, \mathcal{E}, \bm{A})$, where $\mathcal{V} = \{v_1, \cdots, v_N\}$ is a set of $N$ nodes ($|V|=N)$, $\mathcal{E} \subseteq \mathcal{V} \times \mathcal{V}$ is a set of edges, and $\bm{A}$ is the adjacency matrix of network $\mathcal{G}$. Here, $N$ denotes the number of nodes in the graph.
\end{mydef}

% \begin{mydef}[Spatial Region]
% We partition a city into $I \times J$ disjoint grids with the geographical coordinates and represent each grid as a spatial region $r_{i,j}(i \in [1, \cdots, I], j \in [1, \cdots, J])$. To be consistent with the road network data, we define $N$ as the total number of grids, that is, $N = I \times J$. In this way, we can regard the spatial regions as an undirected graph $\mathcal{G}$ with $N$ nodes.
% \end{mydef}

% In this paper, we use the traffic flow of nodes in a traffic system as the observation value.
\begin{mydef}[Traffic Flow Tensor]
We use ${\bm{X}_t} \in {\mathbb{R}^{N \times C}}$ to denote the traffic flow at time $t$ of $N$ nodes in the road network, where $C$ is the dimension of the traffic flow. For example, $C=2$ when the data includes inflow and outflow. We use ${\bm{\mathcal{X}}} = ({{\bm{X}_1}, {\bm{X}_2}, \cdots, {\bm{X}_T}}) \in \mathbb{R}^{T \times N \times C}$ to denote the traffic flow tensor of all nodes at total $T$ time slices.
\end{mydef}
% Here for the graph-based dataset, $C=1$ represents the traffic flow, and for the grid-based dataset, $C=2$ represents the traffic inflow and outflow.

\subsection{Problem Formalization}
Traffic flow prediction aims to predict the traffic flow of a traffic system in the future time given the historical observations. Formally, given the traffic flow tensor $\bm{\mathcal{X}}$ observed on a traffic system, our goal is to learn a mapping function $f$ from the previous $T$ steps' flow observation value to predict future $T'$ steps' traffic flow,
\begin{equation}\label{eq:problem_def}
[\bm{X}_{(t-T+1)}, \cdots, \bm{X}_t; \mathcal{G}] \stackrel{f}{\longrightarrow} [{\bm{X}}_{(t+1)}, \cdots, {\bm{X}}_{(t+T')}].
\end{equation}

% \subsection{Transformer}
% Vanilla Transformer~\cite{transformer} is composed of a stack of Transformer layers. Each Transformer layer has two sub-layers: a multi-head self-attention (${\rm MHA}$) layer and a position-wise fully connected feed-forward network ($\rm  FNN$) layer. Let $\bm{I} \in \mathbb{R}^{T\times {d}}$ denote the input of self-attention module. The multi-head self-attention layer with total $H$ heads simultaneously transforms $\bm{I}$ into $H$ distinct query matrices ${\bm{Q}_h} = \bm{I}\bm{W}_h^Q$, key matrices ${\bm{K}_h}=\bm{I}\bm{W}_h^K$ and value matrices ${\bm{V}_h} = \bm{I}\bm{W}_h^V$. Here $\bm{W}_h^Q, \bm{W}_h^K, \bm{W}_h^V \in \mathbb{R} ^ {{d} \times d'}$ are learnable parameters, and $d' = d / H$. Then the 
% multi-head self attention is calculated as follow:
% \begin{equation} \label{eq:transformer}
% \begin{split}
% & A_h=SA(\bm{Q}_h,\bm{K}_h,\bm{V}_h) = {\rm softmax}\left(\frac{{\bm{Q}_h\bm{K}_h^{\top}}}{{\sqrt{d'}}}\right)\bm{V}_h \\
% & {\rm MHA}(\bm{Q},\bm{K},\bm{V}) = (A_1 ||, \cdots, || A_{H}) \bm{W}^O,
% \end{split}
% \end{equation}
% where $||$ represents concatenation, $AH_h$ is the output of the $h$-th head and all heads are concatenated and projected by ${\bm{W}^O} \in \mathbb{R}^{H{d_v} \times {d}}$ to obtain the final output. After this, a $\rm  FNN$ layer with two linear transformations and a $\rm ReLU$ activation in between is stacked.

\section{Methods}

\begin{figure}[t]
    \centering
    \includegraphics[width=0.9\columnwidth]{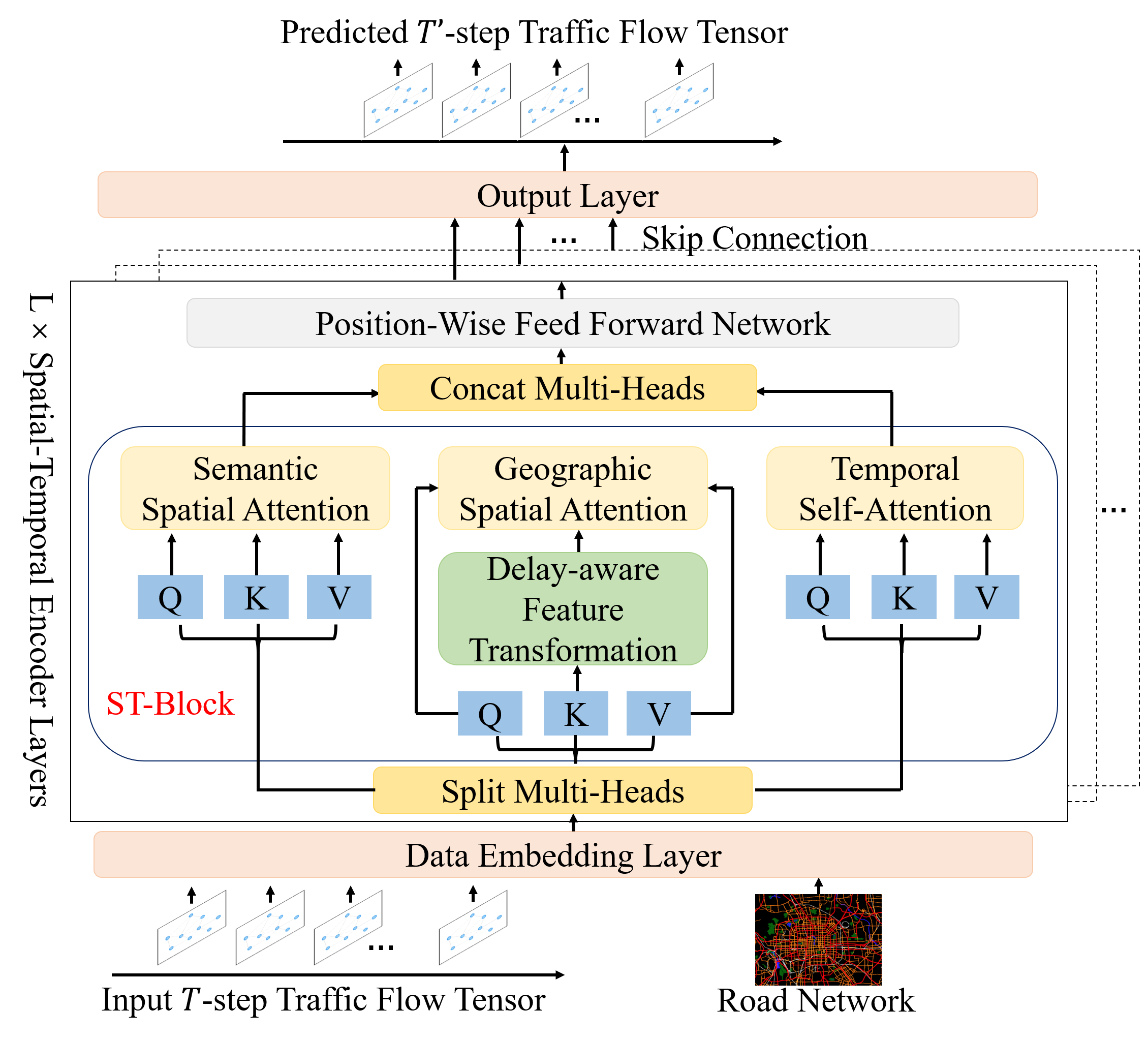}
    \caption{The Overall Framework of \name.}
    \label{fig:framework}
\end{figure}

% \begin{figure}[h]
%     \centering
%     \includegraphics[width=0.9\columnwidth]{figure/overall.png}
%     \caption{The Overall Framework of \name.}
%     \label{fig:framework}
% \end{figure}

% \begin{figure}[h]
%     \centering
%     \includegraphics[width=0.9\columnwidth]{figure/overall3.png}
%     \caption{The Overall Framework of \name.}
%     \label{fig:framework}
% \end{figure}

% \subsection{The Model Framework}
% We present the framework of \name in Fig.~\ref{fig:framework}. It consists of a data embedding layer, stacked $L$ spatial-temporal encoder layers, and an output layer. The data embedding layer projects the raw input $\bm{X} \in \mathbb{R}^{T \times N \times C}$ into a high-dimensional representation $\bm{X}_{emb} \in \mathbb{R}^{T \times N \times d}$. $d$ is the hidden dimension shared by all the stacked encoder layers. We denote the input of the $l$-th encoder layer as $\bm{X}^{(l)}$, where $l \in [1, \cdots, L]$. The output of the ${(l-1)}$-th layer is the input of the $l$-th layer and $\bm{X}^{(1)} = \bm{X}_{emb}$. Each spatial-temporal encoder layer has a skip connection to the output layer. Finally, the output layer projects the hidden features to the desired dimension to get the prediction of the model.

Fig.~\ref{fig:framework} shows the framework of \name, which consists of a data embedding layer, stacked $L$ spatial-temporal encoder layers, and an output layer. We describe each module below in detail.

\subsection{Data Embedding Layer}
The data embedding layer converts the input into a high-dimensional representation. First, the raw input $\bm{\mathcal{X}}$ is transformed into $\bm{\mathcal{X}}_{data} \in \mathbb{R}^{T \times N \times d}$ through a fully connected layer, $d$ is the embedding dimension. Then, we further design a spatial-temporal embedding mechanism to incorporate the necessary knowledge into the model, including the spatial graph Laplacian embedding to encode the road network structure and the temporal periodic embedding to model the periodicity of traffic flow.

% where nodes which are far apart in the graph should have different positional features whereas nodes which are nearby have similar positional features.

To represent the structure of the road network, we use the graph Laplacian eigenvectors~\cite{lap}, which better describe the distance between nodes on the graph. First, we obtain the normalized Laplacian matrix by $\bm{\Delta} = \bm{I} - \bm{D}^{-1/2} \bm{A} \bm{D}^{-1/2}$, where $\bm{A}$ is the adjacency matrix, $\bm{D}$ is the degree matrix, and $\bm{I}$ is the identity matrix. Then, we perform the eigenvalue decomposition $\bm{\Delta} = \bm{U}^{\top}\bm{\Lambda}\bm{U}$ to obtain the eigenvalue matrix $\bm{\Lambda}$ and the eigenvector matrix $\bm{U}$. We use a linear projection on the $k$ smallest nontrivial eigenvectors to generate the spatial graph Laplacian embedding $\bm{X}_{spe} \in \mathbb{R}^{N \times d}$. Laplacian eigenvectors embed the graph in Euclidean space and preserve the global graph structure information~\cite{lap2}.

% Laplacian eigenvectors are hybrid positional and structural encodings that describe the distance between nodes in the graph and preserve the global graph structure. We obtain the Laplacian eigenvectors via the factorization of the graph Laplacian matrix $\bm{\Delta}$:
% \begin{equation}
%     \bm{\Delta} = \bm{I} - \bm{D}^{-1/2} \bm{A} \bm{D}^{-1/2} = \bm{U}^{\top}\bm{\Lambda}\bm{U},
% \end{equation}
% where $\bm{A}$ is the adjacency matrix of graph $\mathcal{G}$, $\bm{D}$ is the degree matrix, $\bm{I}$ is the identity matrix, $\Lambda$ and $\bm{U}$ correspond to the eigenvalues and eigenvectors, respectively. We use a linear projection on $k$ smallest non-trivial eigenvectors to get the spatial embeddings $\bm{X}_{spe} \in \mathbb{R}^{N \times d}$.

In addition, urban traffic flow, influenced by people's travel patterns and lifestyle, has an obvious periodicity, such as morning and evening peak hours. Therefore, we introduce two embeddings to cover the weekly and daily periodicity, respectively, denoted as $\bm{t}_{w(t)}, \bm{t}_{d(t)} \in \mathbb{R}^{d}$. Here ${w{(t)}}$ and ${d{(t)}}$ are functions that transform time ${t}$ into the week index (1 to 7) and minute index (1 to 1440), respectively. The temporal periodic embeddings $\bm{X}_w,\bm{X}_d \in \mathbb{R}^{T \times d}$ are obtained by concatenating the embeddings of all $T$ time slices.

% The above Laplacian eigenvectors encoding can also be considered as spatial position encoding.
% via cosine and sinusoidal functions 
Following the original Transformer~\cite{transformer}, we also employ a temporal position encoding $\bm{X}_{tpe} \in \mathbb{R}^{T \times d}$ to introduce position information of the input sequence.
% as:
% \begin{equation}\label{eq:pe}
% \begin{split}
% & \bm{X}_{tpe}(t, 2i) = \sin(t/10000^{2i/{d}}),\\
% & \bm{X}_{tpe}(t, 2i+1) = \cos(t/10000^{2i/{d}}),
% \end{split}
% \end{equation}
% where $t \in \{0, 1, \cdots, T-1\}$ is the relative index of each element in current input and $i$ is the dimension. 

Finally, we get the output of the data embedding layer by simply summing the above embedding vectors as:
\begin{equation}
    \bm{\mathcal{X}}_{emb} = \bm{\mathcal{X}}_{data} + \bm{X}_{spe} + \bm{X}_{w} + \bm{X}_{d} + \bm{X}_{tpe}.
\end{equation}
$\bm{\mathcal{X}}_{emb}$ will be fed into the following spatial-temporal encoders, and we use $\bm{\mathcal{X}}$ to replace $\bm{\mathcal{X}}_{emb}$ for convenience.

\subsection{Spatial-Temporal Encoder Layer}

% Modeling complex and dynamic spatial-temporal dependencies of data is a key challenge for traffic prediction due to dynamic changes in traffic conditions (\eg peak hours, weekends, and traffic accidents). 

We design a spatial-temporal encoder layer based on the self-attention mechanism to model the complex and dynamic spatial-temporal dependencies. The core of the encoder layer includes three components. The first is a spatial self-attention module consisting of a geographic spatial self-attention module and a semantic spatial self-attention module to capture the short-range and long-range dynamic spatial dependencies simultaneously. The second is a delay-aware feature transformation module that extends the geographic spatial self-attention module to explicitly model the time delay in spatial information propagation. Moreover, the third is a temporal self-attention module that captures the dynamic and long-range temporal patterns.

% In the spatial-temporal self-attention block, we carefully design three self-attention mechanisms, including \emph{Geographic Spatial Self-Attention Module}, \emph{Semantic Spatial Self-Attention Module}, and \emph{Temporal Self-Attention Module}. We will describe them in detail. 

To formulate self-attention operations, we use the following slice notation. For a tensor $\bm{\mathcal{X}} \in \mathbb{R}^{T\times N\times D}$, the $t$ slice is the matrix $\bm{X}_{t::} \in \mathbb{R}^{N\times D}$ and the $n$ slice is $\bm{X}_{:n:} \in \mathbb{R}^{T\times D}$.

% In addition, we factorize the multi-head self-attention operation and adopt the Spatial Self-Attention (\ssa) and Temporal Self-Attention (\tsa) over the spatial and temporal dimensions with different heads, namely \textit{Spatial Attention Heads} (\sah) and \textit{Temporal Attention Heads} (\tah). The results of these heads are concatenated and projected again to obtain the final values, allowing the model to integrate spatial and temporal information simultaneously. Formally, our spatial-temporal self-attention is defined as:
% \begin{equation}
%     \mathrm{STAttn}=\oplus (\mathrm{SAH}_{1\dots h_s},\mathrm{TAH}_{1\dots h_t})\bm{W}^O,
% \end{equation}
% where $\oplus$ represents concatenation, $h_s$, $h_t$ are the numbers of \sah and \tah respectively, and $\bm{W}^O \in \mathbb{R}^{d \times d}$ is a learnable projection matrix. In this work, we set the dimension of query, key and value matrix of the self-attention operation to the same $d' = d / (h_s + h_t)$.

\paratitle{Spatial Self-Attention (\ssa).} We design a \emph{Spatial Self-Attention} module to capture dynamic spatial dependencies in traffic data. Formally, at time $t$, we first obtain the query, key, and value matrices of self-attention operations as:
% \begin{equation}\label{eq:sqkv}
%     \bm{Q}^{(S)}_{t::} = \bm{X}_{t::} \bm{W}^{SQ},
%     \bm{K}^{(S)}_{t::} = \bm{X}_{t::} \bm{W}^{SK},
%     \bm{V}^{(S)}_{t::} = \bm{X}_{t::} \bm{W}^{SV}, 
% \end{equation}
% \begin{equation}\label{eq:sqkv}
%     \bm{Q}^{(S)}_{t::} = \bm{X}_{t::} \bm{W}^{SQ},
%     \bm{K}^{(S)}_{t::} = \bm{X}_{t::} \bm{W}^{SK},
%     \bm{V}^{(S)}_{t::} = \bm{X}_{t::} \bm{W}^{SV}, 
% \end{equation}
\begin{equation}\label{eq:sqkv}
    \bm{Q}^{(S)}_{t} = \bm{X}_{t::} \bm{W}^{S}_Q,
    \bm{K}^{(S)}_{t} = \bm{X}_{t::} \bm{W}^{S}_K,
    \bm{V}^{(S)}_{t} = \bm{X}_{t::} \bm{W}^{S}_V, 
\end{equation}
% \begin{equation}\label{eq:sqkv}
%     \bm{Q}^{(S)}_{t} = \bm{X}_{t::} \bm{W}^{(S)}_Q,
%     \bm{K}^{(S)}_{t} = \bm{X}_{t::} \bm{W}^{(S)}_K,
%     \bm{V}^{(S)}_{t} = \bm{X}_{t::} \bm{W}^{(S)}_V, 
% \end{equation}
where $\bm{W}^{S}_Q, \bm{W}^{S}_K, \bm{W}^{S}_V \in \mathbb{R}^{d \times d'}$ are learnable parameters and $d'$ is the dimension of the query, key, and value matrix in this work. Then, we apply self-attention operations in the spatial dimension to model the interactions between nodes and obtain the spatial dependencies (attention scores) among all nodes at time $t$ as:
\begin{equation}\label{eq:sscore}
    \bm{A}^{(S)}_t=\frac{(\bm{Q}^{(S)}_{t})({\bm{K}^{(S)}_{t}})^{\top}}{\sqrt{d'}}.
\end{equation}
It can be seen that the spatial dependencies $\bm{A}^{(S)}_t \in \mathbb{R}^{N \times N}$ between nodes are different in different time slices, \ie \emph{dynamic}. Thus, the \ssa module can be adapted to capture the dynamic spatial dependencies. Finally, we can obtain the output of the spatial self-attention module by multiplying the attention scores with the value matrix as:
% \begin{equation}\label{eq:ssa}
%     \mathrm{SSA}(\bm{\mathcal{Q}}^{(S)},\bm{\mathcal{K}}^{(S)},\bm{\mathcal{V}}^{(S)})=\mathrm{softmax}(\mathcal{A}^{(S)})\bm{\mathcal{V}}^{(S)}.
% \end{equation}
\begin{equation}\label{eq:ssa}
    \mathrm{SSA}(\bm{Q}^{(S)}_{t},\bm{K}^{(S)}_{t},\bm{V}^{(S)}_{t})=\mathrm{softmax}(\bm{A}^{(S)}_t)\bm{V}^{(S)}_{t}.
\end{equation}

% at time $t$, the spatial dependencies $(A_{n,n}^{(S)})_t$ between all nodes of the $h$-th spatial attention head can be calculated as:
% \begin{equation}\label{eq:ssa}
% \begin{split}
%     & (A_{n,n}^{(S)})_t=\frac{(\bm{Q}_{n,d}^{(S)})_t((\bm{K}_{n,d}^{(S)})_t)^{\top}}{\sqrt{d'}}, \\
%     & (\bm{Q}^{(S)}_{n,d})_t = (\bm{X}_{n,d})_t \bm{W}^{SQ}_h, \\
%     & (\bm{K}^{(S)}_{n,d})_t = (\bm{X}_{n,d})_t \bm{W}^{SK}_h, \\
%     & (\bm{V}^{(S)}_{n,d})_t = (\bm{X}_{n,d})_t \bm{W}^{SV}_h, \\
% \end{split}
% \end{equation}
% where $\bm{W}^{SQ}_h, \bm{W}^{SK}_h, \bm{W}^{SV}_h \in \mathbb{R}^{d \times d'}$ are learnable parameters of the $h$-th spatial head. Then, we can obtain the output of the spatial self-attention by multiplying the attention score with the value matrix as:

% \begin{equation}
%     \mathrm{SSA}(\bm{Q}^{(S)},\bm{K}^{(S)},\bm{V}^{(S)})=\mathrm{softmax}(A^{(S)})\bm{V}^{(S)}.
% \end{equation}

% At time $t$, we first obtain the query, key and value matrices through $(\bm{Q}^{(S)}_{n,d})_t = (\bm{X}_{n,d})_t \bm{W}^{SQ}_h, (\bm{K}^{(S)}_{n,d})_t = (\bm{X}_{n,d})_t \bm{W}^{SK}_h, (\bm{V}^{(S)}_{n,d})_t = (\bm{X}_{n,d})_t \bm{W}^{SV}_h$, where $\bm{W}^{SQ}_h, \bm{W}^{SK}_h, \bm{W}^{SV}_h \in \mathbb{R}^{d \times d'}$ are learnable parameters of the $h$-th spatial head.

% This interaction ignores the significant node pairs and is inadequate to capture the complex spatial dependencies.
For the simple spatial self-attention in Eq.~(\ref{eq:ssa}), each node interacts with all nodes, equivalent to treating the spatial graph as a fully connected graph. However, only the interaction between a few node pairs is essential, including nearby node pairs and node pairs that are far away but have similar functions. Therefore, we introduce two graph masking matrices $\bm{M}_{geo}$ and $\bm{M}_{sem}$ to simultaneously capture the \emph{short-range} and \emph{long-range} spatial dependencies in traffic data.

From the short-range view, we define the binary geographic masking matrix $\bm{M}_{geo}$, and only if the distance (\ie hops in the graph) between two nodes is less than a threshold $\lambda$, the weight is 1, and 0 otherwise. In this way, we can mask the attention of node pairs far away from each other. From the long-range view, we compute the similarity of the historical traffic flow between nodes using the Dynamic Time Warping (DTW)~\cite{dtw} algorithm. We select the $K$ nodes with the highest similarity for each node as its semantic neighbors. Then, we construct the binary semantic masking matrix $\bm{M}_{sem}$ by setting the weight between the current node and its semantic neighbors to 1 and 0 otherwise. In this way, we can find distant node pairs that exhibit similar traffic patterns due to similar urban functions.

Based on the two graph masking matrices, we further design two spatial self-attention modules, namely, \emph{Geographic Spatial Self-Attention} (\gssa) and \emph{Semantic Spatial Self-Attention} (\sssa), which can be defined as:
\begin{equation}\label{eq:fullssa}\small
\begin{split}
    & \mathrm{GeoSSA}(\bm{Q}^{(S)}_{t},\bm{K}^{(S)}_{t},\bm{V}^{(S)}_{t})=\mathrm{softmax}(\bm{A}^{(S)}_t\odot \bm{M}_{geo})\bm{V}^{(S)}_{t}, \\
    & \mathrm{SemSSA}(\bm{Q}^{(S)}_{t},\bm{K}^{(S)}_{t},\bm{V}^{(S)}_{t})=\mathrm{softmax}(\bm{A}^{(S)}_t\odot \bm{M}_{sem})\bm{V}^{(S)}_{t}, \\
\end{split}
\end{equation}
% \begin{equation}\label{eq:fullssa}\small
% \begin{split}
%     & \mathrm{GeoSSA}(\bm{Q}^{(S)}_{t::},\bm{K}^{(S)}_{t::},\bm{V}^{(S)}_{t::})=\mathrm{softmax}(\bm{A}^{(S)}_t\odot \bm{M}_{geo})\bm{V}^{(S)}_{t::}, \\
%     & \mathrm{SemSSA}(\bm{Q}^{(S)}_{t::},\bm{K}^{(S)}_{t::},\bm{V}^{(S)}_{t::})=\mathrm{softmax}(\bm{A}^{(S)}_t\odot \bm{M}_{sem})\bm{V}^{(S)}_{t::}, \\
% \end{split}
% \end{equation}
% \begin{equation}\label{eq:fullssa}\small
% \begin{split}
%     & \mathrm{GeoSSA}(\bm{\mathcal{Q}}^{(S)},\bm{\mathcal{K}}^{(S)},\bm{\mathcal{V}}^{(S)})=\mathrm{softmax}(\mathcal{A}^{(S)}\odot \bm{M}_{geo})\bm{\mathcal{V}}^{(S)}, \\
%     & \mathrm{SemSSA}(\bm{\mathcal{Q}}^{(S)},\bm{\mathcal{K}}^{(S)},\bm{\mathcal{V}}^{(S)})=\mathrm{softmax}(\mathcal{A}^{(S)}\odot \bm{M}_{sem})\bm{\mathcal{V}}^{(S)}, \\
% \end{split}
% \end{equation}
where $\odot$ indicates the Hadamard product. In this way, the spatial self-attention module simultaneously incorporates short-range geographic neighborhood and long-range semantic neighborhood information.

% Then, we decompose \textit{Spatial Attention Heads} (\sah) into \textit{Geographic Spatial Attention Heads} (\gsah) and \textit{Semantic Spatial Attention Heads} (\ssah), which highlight the interactions between significant node pairs by introducing the corresponding spatial mask matrices $\bm{M}_{geo}$ and $\bm{M}_{sem}$. The modified masked spatial self-attention is calculated as:
% \begin{equation}
% \begin{split}
%     & \mathrm{GeoSSA}(\bm{Q}^{(S)},\bm{K}^{(S)},\bm{V}^{(S)})=\mathrm{softmax}(A^{(S)}\odot \bm{M}_{geo})\bm{V}^{(S)}, \\
%     & \mathrm{SemSSA}(\bm{Q}^{(S)},\bm{K}^{(S)},\bm{V}^{(S)})=\mathrm{softmax}(A^{(S)}\odot \bm{M}_{sem})\bm{V}^{(S)}, \\
% \end{split}
% \end{equation}
% where $\odot$ indicates the Hadamard product.

\paratitle{Delay-aware Feature Transformation (\dft).}
%In canonical spatial self-attention operation, each node only attends to the information of other nodes at the same time slice, which is similar to the immediate message passing in GNN-based models and neglects the \textit{propagation delay} of traffic conditions.
There exists a \emph{propagation delay} in real-world traffic conditions. For example, when a traffic accident occurs in one region, it may take several minutes to affect traffic conditions in neighboring regions. Therefore, we propose a traffic delay-aware feature transformation module that captures the propagation delay from the short-term historical traffic flow of each node. Then, this module incorporates delay information into the key matrix of the geographic spatial self-attention module to explicitly model the time delay in spatial information propagation. Since traffic data can have multiple dimensions, such as inflow and outflow, here we only present the calculation process of this module using one dimension as an example.

% \begin{figure}[t]
%     \centering
%     \includegraphics[width=0.9\columnwidth]{figure/pattern.png}
%     \caption{\textcolor{blue}{Delay-aware Feature Transformation.}}
%     \label{fig:pattern}
% \end{figure}

\begin{figure}[t]
    \centering
    \includegraphics[width=0.9\columnwidth]{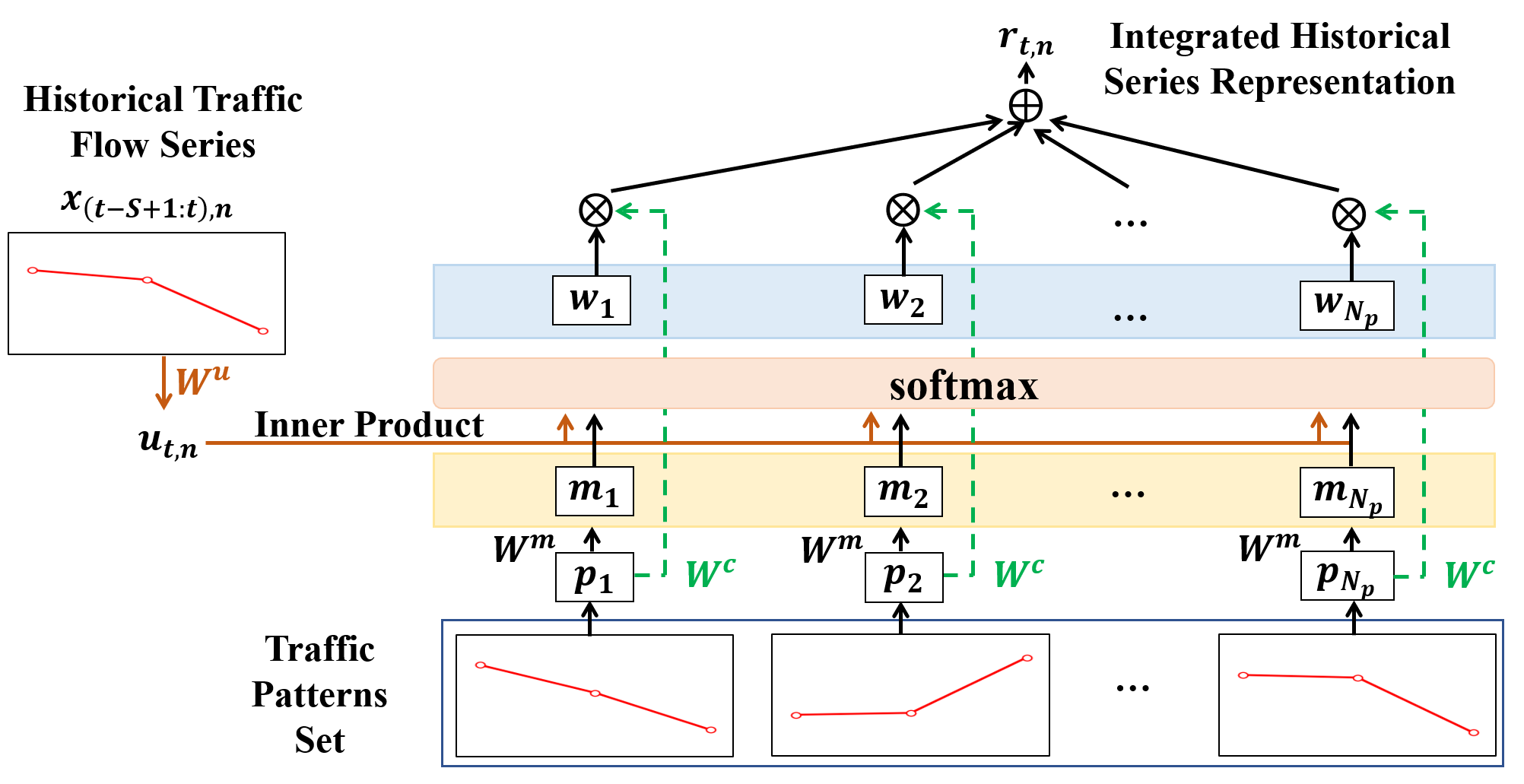}
    \caption{Delay-aware Feature Transformation.}
    \label{fig:pattern}
\end{figure}

% In computing the cluster centroids, the k-Shape method aims to find a centroid most similar to all other time series in this cluster according to the Shape-based distance (SBD).
% , denoted as $\{\bm{x}_{j:j+S-1}\}$
First, we identify a group of representative short-term traffic patterns from historical traffic data. Specifically, we slice the historical traffic data with a sliding window of size $S$ and obtain a set of traffic flow series. Then, we perform k-Shape clustering algorithm~\cite{kshape} on these traffic flow series. The k-Shape algorithm is a time series clustering method that preserves the shape of the time series and is invariant to scaling and shifting. We use the centroid $\bm{p}_i$ of each cluster to represent that cluster, where $\bm{p}_i$ is also a time series of length $S$. Then, we use the set $\mathcal{P}=\{\bm{p}_i | i \in [1,\cdots, N_p] \}$ to represent the clustering results, where $N_p$ is the total number of clusters. We can regard $\mathcal{P}$ as a set of short-term traffic patterns.

%Inspired by the Memory Network~\cite{memory}, 
% ~\footnote{Since traffic data can have multiple dimensions, such as inflow and outflow, we perform separate operations for each dimension here, \ie $\bm{X}_{(t-S:t)::} \in \mathbb{R}^{N \times S}$.}

Similar traffic patterns may have similar effects on neighborhood traffic conditions, especially abnormal traffic patterns such as congestion. Therefore, we compare the historical traffic flow series for each node with the extracted traffic pattern set $\mathcal{P}$ to fuse the information of similar patterns into the historical flow series representation of each node as shown in Fig.~\ref{fig:pattern}. Specifically, given the $S$-step historical traffic flow series of node $n$ from time slice $(t-S+1)$ to $t$, denoted as $\bm{x}_{(t-S+1:t),n}$, we first use the embedding matrix $\bm{W}^{u}$ to obtain a high-dimensional representation $\bm{u}_{t,n}$ as:
\begin{equation}
    \bm{u}_{t,n}=\bm{x}_{(t-S+1:t),n}\bm{W}^{u}.
\end{equation}
Then, we use another embedding matrix $\bm{W}^{m}$ to convert each traffic flow series in the traffic pattern set $\mathcal{P}$ into a memory vector as:
\begin{equation}
    \bm{m}_i = \bm{p}_i\bm{W}^{m}.
\end{equation}
We compare the historical traffic flow representation $\bm{u}_{t,n}$ of node $n$ with the traffic pattern memory vector $\bm{m}_i$ and obtain the similarity vector as:
\begin{equation}
    \bm{w}_i = \mathrm{softmax}(\bm{u}_{t,n}^{\top}\bm{m}_i).
\end{equation}
Then, we perform a weighted sum of the traffic pattern set $\mathcal{P}$ according to the similarity vector $\bm{w}$ to obtain the integrated historical series representation $\bm{r}_{t,n}$ as:
\begin{equation}
    \bm{r}_{t,n} = \sum_{i=1}^{N_p}\bm{w}_i(\bm{p}_i\bm{W}^{c}),
\end{equation}
where $\bm{W}^{c}$ is a learnable parameter matrix. The integrated historical series representation $\bm{r}_{t,n}$ contains the historical traffic flow information from time slice $(t-S+1)$ to $t$ of node $n$. Finally, we use the integrated representation of $N$ nodes, denoted as $\bm{R}_{t}$, to update $\bm{K}_{t}^{(S)}$ in Eq.~(\ref{eq:sscore}) as:
\begin{equation}
  {\tilde{\bm{K}}_{t}^{(S)}}=\bm{K}_{t}^{(S)}+\bm{R}_{t},
\end{equation}
where $\bm{R}_{t}\in \mathbb{R}^{N\times d'}$ is obtained by concatenating all the integrated representation $\bm{r}_{t,n}$ of $N$ nodes and $d'$ is the dimension of the key matrix in this work.

%The short-term pattern representation of all nodes in the time slice $t$, noted as $\bm\mathbb{R}_t$, is obtained by concatenating the pattern representation $\bm\mathbb{R}_{t,n}$ of all $N$ nodes.
% when calculating $\bm{A}^{(S)}_t$ for \gsah in Eq.~(\ref{eq:sscore}), 

In this way, the new key matrix $\tilde{\bm{K}}_{t}^{(S)}$ at time slice $t$ integrates the historical traffic flow information of all nodes from time slice $(t-S+1)$ to $t$. When computing the product of the query matrix and the new key matrix to obtain the spatial dependencies $\bm{A}^{(S)}_t$ at time slice $t$ in Eq.~(\ref{eq:sscore}), the query matrix can take into account the historical traffic conditions of other nodes. This process explicitly models the \emph{time delay} in spatial information propagation. We do not add this module to the semantic spatial self-attention module because the short-term traffic flow of a distant node has little impact on the current node.

%Through the delay-aware feature transformation layer, we can update valuable prior knowledge information into the key matrix of spatial self-attention, solving the time delay problem of a traffic system.

\paratitle{Temporal Self-Attention (\tsa).} There are dependencies (\eg periodic, trending) between traffic conditions in different time slices, and the dependencies vary in different situations. Thus, we employ a \emph{Temporal Self-Attention} module to discover the dynamic temporal patterns. Formally, for node $n$, we first obtain the query, key, and value matrices as: 
% \begin{equation}\label{eq:tqkv}
%     \bm{Q}^{(T)}_{:n:} = \bm{X}_{:n:} \bm{W}^{TQ},
%     \bm{K}^{(T)}_{:n:} = \bm{X}_{:n:} \bm{W}^{TK},
%     \bm{V}^{(T)}_{:n:} = \bm{X}_{:n:} \bm{W}^{TV},
% \end{equation}
% \begin{equation}\label{eq:tqkv}
%     \bm{Q}^{(T)}_{:n:} = \bm{X}_{:n:} \bm{W}^{T}_Q,
%     \bm{K}^{(T)}_{:n:} = \bm{X}_{:n:} \bm{W}^{T}_K,
%     \bm{V}^{(T)}_{:n:} = \bm{X}_{:n:} \bm{W}^{T}_V,
% \end{equation}
\begin{equation}\label{eq:tqkv}
    \bm{Q}^{(T)}_{n} = \bm{X}_{:n:} \bm{W}^{T}_Q,
    \bm{K}^{(T)}_{n} = \bm{X}_{:n:} \bm{W}^{T}_K,
    \bm{V}^{(T)}_{n} = \bm{X}_{:n:} \bm{W}^{T}_V,
\end{equation}
where $\bm{W}^{T}_Q, \bm{W}^{T}_K, \bm{W}^{T}_V \in \mathbb{R}^{d \times d'}$ are learnable parameters. Then, we apply self-attention operations in the temporal dimension and obtain the temporal dependencies between all time slices for node $n$ as:
\begin{equation}\label{eq:tscore}
    \bm{A}^{(T)}_n=\frac{(\bm{Q}^{(T)}_{n})({\bm{K}^{(T)}_{n}}))^{\top}}{\sqrt{d'}}. 
\end{equation}
It can be seen that the temporal self-attention can discover the dynamic temporal patterns in traffic data that are different for different nodes. Moreover, the temporal self-attention has a global receptive to model the long-range temporal dependencies among all time slices. Finally, we can obtain the output of the temporal self-attention module as:
% \begin{equation}\label{eq:tsa}
%     \mathrm{TSA}(\bm{\mathcal{Q}}^{(T)},\bm{\mathcal{K}}^{(T)},\bm{\mathcal{V}}^{(T)})=\mathrm{softmax}(\mathcal{A}^{(T)})\bm{\mathcal{V}}^{(T)}.
% \end{equation}
\begin{equation}\label{eq:tsa}
    \mathrm{TSA}(\bm{Q}^{(T)}_{n},\bm{K}^{(T)}_{n},\bm{V}^{(T)}_{n})=\mathrm{softmax}(\bm{A}^{(T)}_n)\bm{V}^{(T)}_{n}.
\end{equation}
% \begin{equation}\label{eq:tsa}
%     \mathrm{TSA}(\bm{Q}^{(T)}_{:n:},\bm{K}^{(T)}_{:n:},\bm{V}^{(T)}_{:n:})=\mathrm{softmax}(\bm{A}^{(T)}_n)\bm{V}^{(T)}_{:n:}.
% \end{equation}
% Moreover, the self-attention mechanism has a global receptive to model the long-range temporal dependencies. 
% \begin{equation}\label{eq:tsa}
% \begin{split}
%     & (A_{t,t}^{(T)})_n=\frac{(\bm{Q}_{t,d}^{(T)})_n((\bm{K}_{t,d}^{(T)})_n)^{\top}}{\sqrt{d'}}, \\
%     & (\bm{Q}^{(T)}_{:n:} = (\bm{X}_{:n:} \bm{W}^{TQ}_h, \\
%     & (\bm{K}^{(T)}_{:n:} = (\bm{X}_{:n:} \bm{W}^{TK}_h, \\
%     & (\bm{V}^{(T)}_{:n:} = (\bm{X}_{:n:} \bm{W}^{TV}_h, \\
% \end{split}
% \end{equation}
% where $\bm{W}^{TQ}_h, \bm{W}^{TK}_h, \bm{W}^{TV}_h \in \mathbb{R}^{d \times d'}$ are learnable parameters of the $h$-th temporal head. Then, we can obtain the output of the temporal self-attention as:
% \begin{equation}
%     \mathrm{TSA}(\bm{Q}^{(T)},\bm{K}^{(T)},\bm{V}^{(T)})=\mathrm{softmax}(A^{(T)})\bm{V}^{(T)}.
% \end{equation}

\paratitle{Heterogeneous Attention Fusion.} After defining the three types of attention mechanisms, we fuse heterogeneous attention into a multi-head self-attention block to reduce the computational complexity of the model. Specifically, the attention heads include three types, \ie geographic (\gsah), semantic (\ssah), and temporal (\tah) heads, corresponding to the three types of attention mechanisms, respectively. The results of these heads are concatenated and projected to obtain the outputs, allowing the model to integrate spatial and temporal information simultaneously. Formally, the spatial-temporal self-attention block is defined as:
\begin{equation}\label{eq:total}
\begin{split}
    \mathrm{STAttn}=\oplus (\bm{Z}^{geo}_{1\cdots h_{geo}},\bm{Z}^{sem}_{1\cdots h_{sem}}, \bm{Z}^{t}_{1\cdots h_t})\bm{W}^O,
\end{split}
\end{equation}
where $\oplus$ represents concatenation, $\bm{Z}^{geo}, \bm{Z}^{sem}, \bm{Z}^{t}$ are output concatenations and $h_{geo}$, $h_{sem}$, $h_t$ are the numbers of attention heads of \gssa, \sssa and \tsa, respectively, and $\bm{W}^O \in \mathbb{R}^{d \times d}$ is a learnable projection matrix. In this work, we set the dimension $d' = d / (h_{geo} + h_{sem} + h_t)$.

% We will replace the attention heads(GeoSAH,SemSAH,TAH) with the output vectors(z^g,z^s,z^t) and restate Eq.15.

In addition, we employ a position-wise fully connected feed-forward network on the output of the multi-head self-attention block to get the outputs $\bm{\mathcal{X}}_{o} \in \mathbb{R}^{T \times N \times d}$. We also use layer normalization and residual connection here following the original Transformer~\cite{transformer}. 

% we factorize the multi-head attention operation and adopt the Spatial Self-Attention (\ssa) and Temporal Self-Attention (\tsa) over the spatial and temporal dimensions with different heads, namely \textit{Spatial Attention Heads} (\sah) and \textit{Temporal Attention Heads} (\tah). 

\subsection{Output Layer}

%We use $\hat{\bm{Y}} \in \mathbb{R}^{T' \times N \times C}$ to represent the prediction results for $T'$ future time slices of all $N$ nodes.

We use a skip connection, consisting of 1 × 1 convolutions, after each spatial-temporal encoder layer to convert the outputs $\bm{\mathcal{X}}_{o}$ into a skip dimension $\bm{\mathcal{X}}_{sk} \in \mathbb{R}^{T \times N \times {d_{sk}}}$. Here $d_{sk}$ is the skip dimension. Then, we obtain the final hidden state $\bm{\mathcal{X}}_{hid} \in \mathbb{R}^{T \times N \times {d_{sk}}}$ by summing the outputs of each skip connection layer. To make a multi-step prediction, we directly use the output layer to transform the final hidden state $\bm{\mathcal{X}}_{hid}$ to the desired dimension as:
\begin{equation}\label{eq:final}
    \bm{\hat{\mathcal{X}}} = \mathrm{Conv_2}(\mathrm{Conv_1}(\bm{\mathcal{X}}_{hid})),
\end{equation}
where $\bm{\hat{\mathcal{X}}} \in \mathbb{R}^{T' \times N \times C}$ is $T'$ steps' prediction results, $\mathrm{Conv_1}$ and $\mathrm{Conv_2}$ are 1 × 1 convolutions. Here we choose the direct way instead of the recursive manner for multi-step prediction considering cumulative errors and model efficiency.

\section{Experiments}
% In this section, we evaluate the performance of \name by a series of experiments on six real-world datasets. %, which are summarized to answer the following research questions: 

% \begin{itemize}
% \item \textbf{RQ1}: How is the performance of \name compared to other existing methods?
% \item \textbf{RQ2}: How does each sub-module contribute to the model performance?
% \item \textbf{RQ3}: How is \name sensitive to key hyperparameters?
% \item \textbf{RQ4}: Why \name can yield a good performance?
% \item \textbf{RQ5}: How is the computation efficiency of \name?
% \end{itemize}

\begin{table}[t]
  \centering
  \caption{Data Description.}
  \resizebox{\columnwidth}{!}{
    \begin{tabular}{cccccc}
    \toprule
    Datasets & \#Nodes & \#Edges & \#Timesteps & \#Time Interval & Time range \\
    \midrule
    \pfour & 307   & 340   & 16992 & 5min  & 01/01/2018-02/28/2018 \\
    \pseven & 883   & 866   & 28224 & 5min  & 05/01/2017-08/31/2017 \\
    \peight & 170   & 295   & 17856 & 5min  & 07/01/2016-08/31/2016 \\
    \midrule
    \taxi & 75 (15x5) & 484   & 17520 & 30min & 01/01/2014-12/31/2014 \\
    \bike & 270 (15x18) & 1966  & 4416  & 30min & 07/01/2020-09/30/2020 \\
    \bj & 1024 (32x32) & 7812  & 3600  & 60min & 02/01/2015-06/30/2015 \\
    \bottomrule
    \end{tabular}%
    }
  \label{tab:data_detail}%
\end{table}%

\subsection{Datasets}
We verify the performance of \name on six real-world public traffic datasets, including three graph-based highway traffic datasets, \ie \pfour, \pseven, \peight~\cite{STSGCN}, and three grid-based citywide traffic datasets, \ie \taxi~\cite{nyctaxi}, \bike~\cite{libcity}, \bj~\cite{stmetanet}. The graph-based datasets contain only the traffic flow data, and the grid-based datasets contain inflow and outflow data. Details are given in Tab.~\ref{tab:data_detail}. % We set each grid connected to the surrounding eight grids for the grid-based datasets to create the spatial adjacency matrix $\bm{A}$. % See the appendix for more details.

\subsection{Baselines}
We compare \name with the following 17 baselines belonging to four classes. (1) \textit{Models For Grid-based Datasets}: We choose STResNet~\cite{STResNet}, DMVSTNet~\cite{dmvstnet} and DSAN~\cite{DSAN}, which are unsuitable for graph-based datasets. (2) \textit{Time Series Prediction Models}: We choose VAR~\cite{var} and SVR~\cite{svr}. (3) \textit{Graph Neural Network-based Models}: We choose DCRNN~\cite{DCRNN}, STGCN~\cite{STGCN}, GWNET~\cite{GWNET}, MTGNN~\cite{MTGNN}, STSGCN~\cite{STSGCN}, STFGNN~\cite{STFGNN}, STGODE~\cite{STGODE} and STGNCDE~\cite{STG-NCDE}. (4) \textit{Self-attention-based Models}: We choose STTN~\cite{STTN}, GMAN~\cite{GMAN}, TFormer~\cite{TFormer} and ASTGNN~\cite{astgnn}. % More details are given in Appendix.

\subsection{Experimental Settings}
\paratitle{Dataset Processing.}
To be consistent with most modern methods, we split the three graph-based datasets into training, validation, and test sets in a 6:2:2 ratio. In addition, we use the past hour (12 steps) data to predict the traffic flow for the next hour (12 steps), \ie a multi-step prediction. For the grid-based datasets, the split ratio is 7:1:2, and we use the traffic inflow and outflow of the past six steps to predict the next single-step traffic inflow and outflow.
%Before training, we use Z-score normalization on all datasets to standardize the inputs.

% Table generated by Excel2LaTeX from sheet 'Sheet2'
\begin{table}[t]
  \centering
  \caption{Performance on Graph-based Datasets.}
  \resizebox{\columnwidth}{!}{
    \begin{tabular}{c|ccc|ccc|ccc}
    \toprule
    \multirow{2}[4]{*}{Model} & \multicolumn{3}{c|}{\pfour} & \multicolumn{3}{c|}{\pseven} & \multicolumn{3}{c}{\peight} \\
\cmidrule{2-10}          & MAE   & MAPE(\%) & RMSE  & MAE   & MAPE(\%) & RMSE  & MAE   & MAPE(\%) & RMSE \\
    \hline
    VAR   & 23.750  & 18.090  & 36.660  & 101.200  & 39.690  & 155.140  & 22.320  & 14.470  & 33.830  \\
    SVR   & 28.660  & 19.150  & 44.590  & 32.970  & 15.430  & 50.150  & 23.250  & 14.710  & 36.150  \\
    \hline
    DCRNN & 22.737  & 14.751  & 36.575  & 23.634  & 12.281  & 36.514  & 18.185  & 11.235  & 28.176  \\
    STGCN & 21.758  & 13.874  & 34.769  & 22.898  & 11.983  & 35.440  & 17.838  & 11.211  & 27.122  \\
    GWNET & 19.358  & 13.301  & 31.719  & 21.221  & 9.075  & 34.117  & 15.063  & 9.514  & 24.855  \\
    MTGNN & 19.076  & 12.961  & 31.564  & 20.824  & 9.032  & 34.087  & 15.396  & 10.170  & 24.934  \\
    STSGCN & 21.185  & 13.882  & 33.649  & 24.264  & 10.204  & 39.034  & 17.133  & 10.961  & 26.785  \\
    STFGNN & 19.830  & 13.021  & 31.870  & 22.072  & 9.212  & 35.805  & 16.636  & 10.547  & 26.206  \\
    STGODE & 20.849  & 13.781  & 32.825  & 22.976  & 10.142  & 36.190  & 16.819  & 10.623  & 26.240  \\
    STGNCDE & 19.211  & 12.772  & 31.088  & 20.620  & 8.864  & 34.036  & 15.455  & 9.921  & 24.813  \\
    \hline
    STTN  & 19.478  & 13.631  & 31.910  & 21.344  & 9.932  & 34.588  & 15.482  & 10.341  & 24.965  \\
    GMAN  & 19.139  & 13.192  & 31.601  & 20.967  & 9.052  & 34.097  & 15.307  & 10.134  & 24.915  \\
    TFormer & 18.916  & 12.711  & 31.349  & 20.754  & 8.972  & 34.062  & 15.192  & 9.925  & 24.883  \\
    ASTGNN & \underline{18.601}  & \underline{12.630}  & \underline{31.028}  & \underline{20.616}  & \underline{8.861}  & \underline{34.017}  & \underline{14.974}  & \underline{9.489}  & \underline{24.710}  \\
    \hline
    \name & \textbf{18.321 } & \textbf{12.103 } & \textbf{29.965 } & \textbf{19.832 } & \textbf{8.529 } & \textbf{32.870 } & \textbf{13.583 } & \textbf{9.046 } & \textbf{23.505 } \\
    \bottomrule
    \end{tabular}%
	}
  \label{tab:res_graph}%
\end{table}%

% Table generated by Excel2LaTeX from sheet 'Sheet3'
\begin{table*}[t]
  \centering
  \caption{Performance on Grid-based Datasets.}
   \resizebox{\textwidth}{!}{
    \begin{tabular}{c|ccc|ccc|ccc|ccc|ccc|ccc}
    \toprule
    Datasets & \multicolumn{6}{c|}{\taxi}                  & \multicolumn{6}{c|}{\bj}                  & \multicolumn{6}{c}{\bike} \\
    \hline
    Metrics & \multicolumn{3}{c|}{inflow} & \multicolumn{3}{c|}{outflow} & \multicolumn{3}{c|}{inflow} & \multicolumn{3}{c|}{outflow} & \multicolumn{3}{c|}{inflow} & \multicolumn{3}{c}{outflow} \\
    \hline
    Models & MAE   & MAPE(\%) & RMSE  & MAE   & MAPE(\%) & RMSE  & MAE   & MAPE(\%) & RMSE  & MAE   & MAPE(\%) & RMSE  & MAE   & MAPE(\%) & RMSE  & MAE   & MAPE(\%) & RMSE \\
    \hline
    STResNet & 14.492  & 14.543  & 24.050  & 12.798  & 14.368  & 20.633  & 19.636  & 17.831  & 34.890  & 19.616  & 18.502  & 34.597  & 4.767  & 31.382  & 6.703  & 4.627  & 30.571  & 6.559  \\
    DMVSTNet & 14.377  & 14.314  & 23.734  & 12.566  & 14.318  & 20.409  & 19.599  & 17.683  & 34.478  & 19.531  & 17.621  & 34.303  & 4.687  & 32.113  & 6.635  & 4.594  & 31.313  & 6.455  \\
    DSAN  & 14.287  & 14.208  & 23.585  & 12.462  & 14.272  & 20.294  & 19.384  & 17.465  & 34.314  & 19.290  & 17.379  & 34.267  & 4.612  & 31.621  & 6.695  & 4.495  & 31.256  & 6.367  \\
    \hline
    DCRNN & 14.421  & 14.353  & 23.876  & 12.828  & 14.344  & 20.067  & 22.121  & 17.750  & 38.654  & 21.755  & 17.382  & 38.168  & 4.236  & 31.264  & 5.992  & 4.211  & 30.822  & 5.824  \\
    STGCN & 14.377  & 14.217  & 23.860  & 12.547  & 14.095  & 19.962  & 21.373  & 17.539  & 38.052  & 20.913  & 16.984  & 37.619  & 4.212  & 31.224  & 5.954  & 4.148  & 30.782  & 5.779  \\
    GWNET & 14.310  & 14.198  & 23.799  & 12.282  & 13.685  & 19.616  & 19.556  & 17.187  & 36.159  & 19.550  & 15.933  & 36.198  & 4.151  & 31.153  & 5.917  & 4.101  & 30.690  & 5.694  \\
    MTGNN & 14.194  & 13.984  & 23.663  & 12.272  & 13.652  & 19.563  & 18.982  & 17.056  & 35.386  & 18.929  & 15.762  & 35.992  & 4.112  & 31.148  & 5.807  & 4.086  & 30.561  & 5.669  \\
    STSGCN & 15.604  & 15.203  & 26.191  & 13.233  & 14.698  & 21.653  & 23.825  & 18.547  & 41.188  & 24.287  & 19.041  & 42.255  & 4.256  & 32.991  & 5.941  & 4.265  & 32.612  & 5.879  \\
    STFGNN & 15.336  & 14.869  & 26.112  & 13.178  & 14.584  & 21.627  & 22.144  & 18.094  & 40.071  & 22.876  & 18.987  & 41.037  & 4.234  & 32.222  & 5.933  & 4.264  & 32.321  & 5.875  \\
    STGODE & 14.621  & 14.793  & 25.444  & 12.834  & 14.398  & 20.205  & 21.515  & 17.579  & 38.215  & 22.703  & 18.509  & 40.282  & 4.169  & 31.165  & 5.921  & 4.125  & 30.726  & 5.698  \\
    STGNCDE & 14.281  & 14.171  & 23.742  & 12.276  & 13.681  & 19.608  & 19.347  & 17.134  & 36.093  & 19.230  & 15.873  & 36.143  & 4.123  & 31.151  & 5.913  & 4.094  & 30.595  & 5.678  \\
    \hline
    STTN  & 14.359  & 14.206  & 23.841  & 12.373  & 13.762  & 19.827  & 20.583  & 17.327  & 37.220  & 20.443  & 15.992  & 37.067  & 4.160  & 31.208  & 5.932  & 4.118  & 30.704  & 5.723  \\
    GMAN  & 14.267  & 14.114  & 23.728  & 12.273  & 13.672  & 19.594  & 19.244  & 17.110  & 35.986  & 18.964  & 15.788  & 36.120  & 4.115  & 31.150  & 5.910  & 4.090  & 30.662  & 5.675  \\
    TFormer & 13.995  & 13.912  & 23.487  & 12.211  & 13.611  & 19.522  & 18.823  & 16.910  & 34.470  & 18.883  & 15.674  & 35.219  & 4.071  & 31.141  & 5.878  & 4.037  & 30.647  & 5.638  \\
    ASTGNN & \underline{13.844}  & \underline{13.692}  & \underline{23.177}  & \underline{12.112}  & \underline{13.602}  & \underline{19.201}  & \underline{18.798}  & \underline{16.101}  & \underline{33.870}  & \underline{18.790}  & \underline{15.584}  & \underline{33.998}  & \underline{4.068}  & \underline{31.131}  & \underline{5.818}  & \underline{3.981}  & \underline{30.617}  & \underline{5.609}  \\
    \hline
    \name & \textbf{13.152 } & \textbf{12.743 } & \textbf{21.957 } & \textbf{11.575 } & \textbf{12.820 } & \textbf{18.394 } & \textbf{17.832 } & \textbf{14.711 } & \textbf{31.606 } & \textbf{17.743 } & \textbf{14.649 } & \textbf{31.501 } & \textbf{3.950 } & \textbf{30.214 } & \textbf{5.559 } & \textbf{3.837 } & \textbf{29.914 } & \textbf{5.402 } \\
    \bottomrule
    \end{tabular}%
    }
  \label{tab:res_grid}%
\end{table*}%

\paratitle{Model Settings.}
All experiments are conducted on a machine with the NVIDIA GeForce 3090 GPU and 128GB memory. We implement \name~\footnote{\url{https://github.com/BUAABIGSCity/PDFormer}} with Ubuntu 18.04, PyTorch 1.10.1, and Python 3.9.7. The hidden dimension $d$ is searched over \{16, 32, 64, 128\} and the depth of encoder layers $L$ is searched over \{2, 4, 6, 8\}. The optimal model is determined based on the performance in the validation set. We train our model using AdamW optimizer~\cite{adam} with a learning rate of 0.001. The batch size is 16, and the training epoch is 200. 
%Following previous work~\cite{STSGCN, STFGNN}, we use the Huber loss~\cite{huber1992robust} as the training loss for both our model and baselines as it is less sensitive to outliers than the squared error loss. See the appendix for more detailed settings.

\paratitle{Evaluation Metrics.} We use three metrics in the experiments: (1) Mean Absolute Error (MAE), (2) Mean Absolute Percentage Error (MAPE), and (3) Root Mean Squared Error (RMSE). Missing values are excluded when calculating these metrics. When we test the models on grid-based datasets, we filter the samples with flow values below 10, consistent with~\cite{dmvstnet}. Since the flow of CHIBike is lower than others, the filter threshold is 5. We repeated all experiments ten times and reported the average results.

% \vspace{-0.2cm}
\subsection{Performance Comparison}

% The results with standard deviation are given in the appendix for space limitation. 
The comparison results with baselines on graph-based and grid-based datasets are shown in Tab.~\ref{tab:res_graph} and Tab.~\ref{tab:res_grid}, respectively. The bold results are the best, and the underlined results are the second best. Based on these two tables, we can make the following observations. (1) Spatial-temporal deep learning models perform better than traditional time series prediction models such as VAR because the latter ignores the spatial dependencies in the traffic data. (2) Our \name significantly outperforms all baselines in terms of all metrics over all datasets according to Student's t-test at level 0.01. Compared to the second best method, \name achieves an average improvement of 4.58\%, 5.00\%, 4.79\% for MAE/MAPE/RMSE. (3) Among the GNN-based models, MTGNN and STGNCDE lead to competitive performance. Compared to these GNN-based models, whose message passing is immediate, \name achieves better performance because it considers the time delay in spatial information propagation. (4) As for the self-attention-based models, ASTGNN is the best baseline, which combines GCN and the self-attention module to aggregate neighbor information. Compared with ASTGNN, \name simultaneously captures short- and long-range spatial dependencies via two masking matrices and achieves good performance. (5) Although STResNet, DMVSTNet, and DSAN are designed for grid-based data, they need to extract features from different time periods, which are excluded here, resulting in relatively poor performance.

\subsection{Ablation Study}
To further investigate the effectiveness of different parts in \name, we compare \name with the following variants. (1) \textit{w/ GCN}: this variant replaces spatial self-attention (\ssa) with Graph Convolutional Network (GCN)~\cite{gcn}, which cannot capture the dynamic and long-range spatial dependencies. (2) \textit{w/o Mask}: this variant removes two masking matrices $\bm{M}_{geo}$ and $\bm{M}_{sem}$, which means each node attends to all nodes. (3) \textit{w/o GeoSAH}: this variant removes \gsah. (4) \textit{w/o SemSAH}: this variant removes \ssah. (5) \textit{w/o Delay}: this variant removes the delay-aware feature transformation module, which accounts for the spatial information propagation delay.

\begin{figure}[t]
    \centering
    \includegraphics[width=0.9\columnwidth]{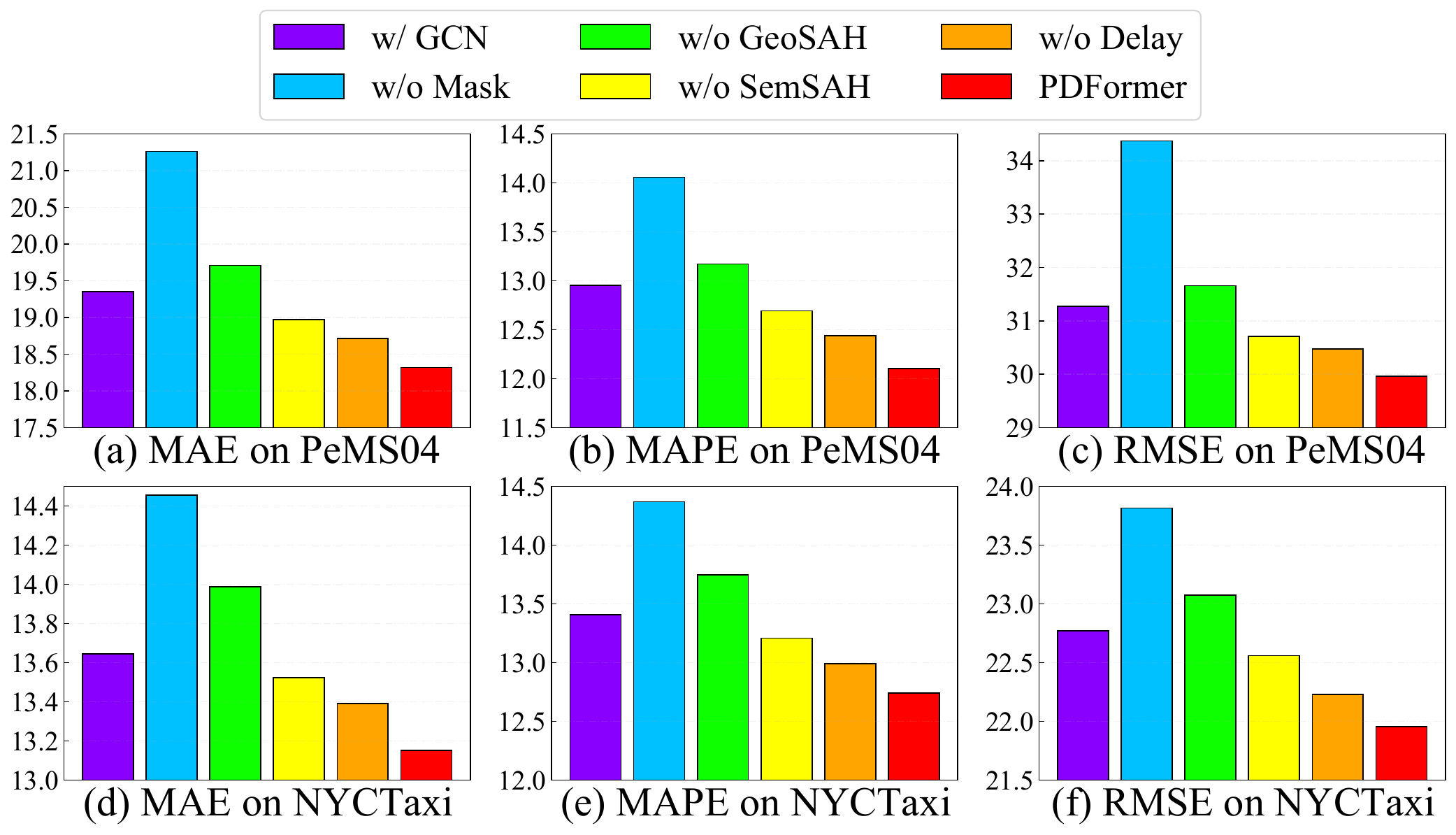}
    \caption{Ablation Study on \pfour and \taxi inflow.}
    \label{fig:abla}
\end{figure}

Fig.~\ref{fig:abla} shows the comparison of these variants on the \pfour and \taxi datasets. For the \taxi dataset, only the results for inflow are reported since the results for outflow are similar. Based on the results, we can conclude the following: (1) The results show the superiority of \ssa over GCN in capturing dynamic and long-range spatial dependencies. (2) \name leads to a large performance improvement over \textit{w/o Mask}, highlighting the value of using the mask matrices to identify the significant node pairs. In addition, \textit{w/o SemSAH} and \textit{w/o GeoSAH} perform worse than \name, indicating that both local and global spatial dependencies are significant for traffic prediction. (3) \textit{w/o Delay} performs worse than \name because this variant ignores the spatial propagation delay between nodes but considers the spatial message passing as immediate.

% \subsection{Parameter Sensitivity}

% \begin{figure}[h]
%     \centering
%     \includegraphics[width=0.9\columnwidth]{figure/para.pdf}
%     \caption{Parameter Analysis on \pfour and \taxi.}
%     \label{fig:para}
% \end{figure}

% We further conduct the parameter sensitivity analysis of \name on \pfour and \taxi datasets. There are three core hyper-parameters to be manually set, consisting of the embedding dimension $d$, the depth of encoder layers $L$, and the ratio of spatial-temporal heads, \ie \emph{SAH} and \emph{TAH}. Fig.~\ref{fig:para} reports the results. We observe that the model performance improves with the embedding dimension or model depths become larger. However, when they are too large, the performance degradation due to over-fitting. Besides, the ratio of \emph{SAH} and \emph{TAH} can flexibly control the proportion of learned spatial and temporal information. \name is relatively stable when varying this parameter except for 0:8, because spatial information is not considered in this case.

\subsection{Case Study}
In this section, we analyze the dynamic spatial-temporal attention weight map learned by the spatial-temporal encoder of \name to improve its interpretability and demonstrate the effectiveness of focusing on short- and long-range spatial dependencies simultaneously.

\begin{figure}[t]
    \centering
    \includegraphics[width=0.9\columnwidth]{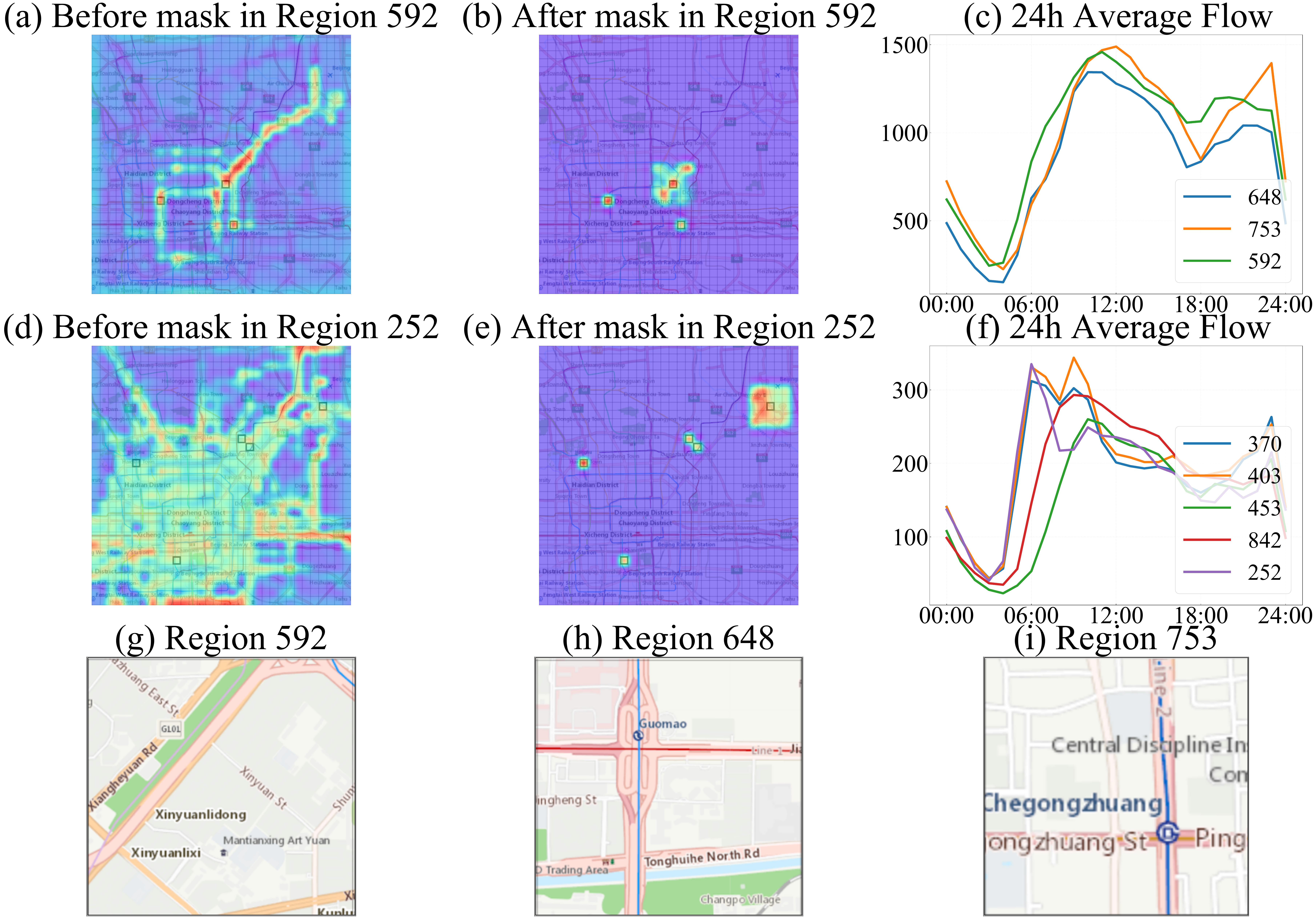}
    \caption{Case Study of Attention Map.}
    \label{fig:case1}
\end{figure}

We compare and visualize the attention map in two cases, \ie with or without the two spatial mask matrices $\bm{M}_{geo}$ and $\bm{M}_{sem}$. Here, for simplicity, we merge the attention map of \gsah and \ssah. As shown in Fig.~\ref{fig:case1}(a),(d), without the mask matrices, the model focuses on the major urban ring roads (or highways) with high traffic volume, or the attention distribution is diffuse, and almost the entire city shares the model's attention. However, low-traffic locations should focus on locations with similar patterns rather than hot locations. Moreover, locations that are too far away have little impact on the current location. The model performance will weaken if it focuses on all locations diffusely. Instead, when $\bm{M}_{geo}$ and $\bm{M}_{sem}$ are introduced, attention focuses on surrounding locations and distant similar-pattern locations as shown in Fig.~\ref{fig:case1}(b),(e).

% \begin{figure}[t]
%     \centering
%     \includegraphics[width=0.9\columnwidth]{figure/case3_old.pdf}
%     \caption{Case Study of Attention Map.}
%     \label{fig:case1}
% \end{figure}

% \begin{figure}[t]
%     \centering
%     \includegraphics[width=0.9\columnwidth]{figure/case3_6x6.pdf}
%     \caption{Case Study of Attention Map.}
%     \label{fig:case1}
% \end{figure}

Let us take Region 592 in Fig.~\ref{fig:case1}(b) as an example. Highway S12 passes through this region, so the traffic volume is always high. In addition to the regions located upstream and downstream of the highway, region 592 also focuses on regions 648 and 753. From Fig.~\ref{fig:case1}(c), we can see that these two regions have similar historical traffic volumes as 592. Besides, from Fig.~\ref{fig:case1}(h)(i), these two regions are located near the Guomao Interchange and Beijing Second Ring Road, respectively, which are similar to 592 in their cities function as major traffic hubs. In another case, region 252 has low traffic volume, but we can observe similar patterns from regions 370, 403, 453, and 842 that region 252 focused on, \ie similar functional and historical traffic variations.

This case study shows that after introducing the spatial mask matrices, \name not only considers the short-range spatial dependencies but also identifies the global functional area to capture the long-range spatial dependencies. The above ablation experiments also quantitatively show that the model performance drops sharply after removing the mask matrices, which supports the idea presented here.

\subsection{Model Efficiency Study}
Due to the better performance of the attention-based models, we compare the computational cost of \name with other self-attention-based baselines on the \pfour and the \taxi datasets. Tab.~\ref{tab:time_cost} reports the average training and inference time per epoch. We find that \name achieves competitive computational efficiency in both short- and long-term traffic prediction. Compared to the best performing baseline ASTGNN on \pfour, \name reduces the training and inference time of over 35\% and 80\%, respectively. GMAN and ASTGNN retain a time-consuming encoder-decoder structure, which is replaced by a forward procedure in STTN and \name. TFormer performs only spatial attention, resulting in less computation time.

\section{Related Work}

\subsection{Deep Learning for Traffic Prediction}
% In recent years, more and more researchers have employed deep learning models to solve traffic prediction problems. Early on, convolutional neural networks (CNNs) were applied to grid-based traffic data to capture spatial dependencies in the data~\cite{STResNet, DSAN}. Later, thanks to the powerful ability to model graph data, graph neural networks (GNNs) were widely used for traffic prediction. GWNET~\cite{GWNET} employed a graph convolution on an adaptive spatial matrix. STFGNN~\cite{STFGNN} captured spatial-temporal dependencies synchronously through a spatial-temporal fusion module. Recently, the attention mechanism has become increasingly popular due to its effectiveness in modeling the dynamic dependencies in traffic data. GMAN~\cite{GMAN} proposed spatial-temporal attention mechanism and designed a gated fusion module. ASTGNN~\cite{astgnn} proposed a dynamic graph convolution module with a self-attention mechanism to capture spatial dependencies. Unlike these work, our proposed \name not only considers the dynamic and long-range spatial dependencies through a self-attention mechanism but also incorporates the time delay in spatial propagation in the traffic system through a delay-aware feature transformation layer.

In recent years, more and more researchers have employed deep learning models to solve traffic prediction problems. Early on, convolutional neural networks (CNNs) were applied to grid-based traffic data to capture spatial dependencies in the data~\cite{STResNet, DSAN}. Later, thanks to the powerful ability to model graph data, graph neural networks (GNNs) were widely used for traffic prediction~\cite{mrange,agcrn,stgdn,fc-gaga, ccrnn,dmstgcn,stden,msdr}. Recently, the attention mechanism has become increasingly popular due to its effectiveness in modeling the dynamic dependencies in traffic data~\cite{astgcn,stgnn,stconvlstm,astgnn,MGT}. Unlike these work, our proposed \name not only considers the dynamic and long-range spatial dependencies through a self-attention mechanism but also incorporates the time delay in spatial propagation through a delay-aware feature transformation layer.

\subsection{Transformer}
% Transformer~\cite{transformer} is a network architecture based entirely on self-attention mechanisms. An important advantage of the self-attention mechanism is its global modeling ability and less inductive bias. The self attention mechanism has a global receptive. Theoretically, each position of the input can attend to all the remaining positions in the input. In recent years, Transformer has been widely used in many domains, such as Natural Language Processing (NLP)~\cite{devlin2018bert},  Computer Vision (CV)~\cite{image, swin} and Graph Representation Learning~\cite{graphtrans, Graphormer}.

% An important advantage of the self-attention mechanism is its global modeling ability and less inductive bias.

Transformer~\cite{transformer} is a network architecture based entirely on self-attention mechanisms. Transformer has been proven effective in multiple natural language processing (NLP) tasks. In addition, large-scale Transformer-based pre-trained models such as BERT~\cite{devlin2018bert} have achieved great success in the NLP community. Recently, Vision Transformers have attracted the attention of researchers, and many variants have shown promising results on computer vision tasks~\cite{image, swin}. In addition, the Transformer architecture performs well in representation learning, which has been demonstrated in recent studies~\cite{graphbert, rapt, jiang2023start}.% For example, Graphormer~\cite{Graphormer} introduces effective structural encoding and achieves state-of-the-art performance on several graph-level prediction tasks.

% Tab. generated by Excel2LaTeX from sheet 'Sheet4'
\begin{table}[t]
  \centering
  \caption{Training and inference time per epoch comparison between self-attention-based models. (Unit: seconds)}
    \resizebox{0.8\columnwidth}{!}{
    \begin{tabular}{c|c|c|c|c}
    \toprule
    Dataset & \multicolumn{2}{c|}{\pfour} & \multicolumn{2}{c}{\taxi} \\
    \hline
    Model & Training & Inference & Training & Inference \\
    \hline
    GMAN  & 501.578 & 38.844 & 130.672 & 4.256 \\
    ASTGNN & 208.724 & 52.016 & 119.092 & 4.601 \\
    %STGFormer & 110.293 & 9.483 & 58.892 & 2.098 \\
    \name & 133.871 & 8.120 & 85.305 & 2.734 \\
    STTN  & 100.398 & 12.596 & 68.036 & 2.650 \\
    TFormer & 71.099 & 7.156 & 76.169 & 2.575 \\
    \bottomrule
    \end{tabular}%
    }
  \label{tab:time_cost}%
\end{table}%

\section{Conclusion}
In this work, we proposed a novel \name model with spatial-temporal self-attention for traffic flow prediction. Specifically, we developed a spatial self-attention module that captures the dynamic and long-range spatial dependencies and a temporal self-attention module that discovers the dynamic temporal patterns in the traffic data. We further designed a delay-aware feature transformation module to explicitly model the time delay in spatial information propagation. We conducted extensive experiments on six real-world datasets to demonstrate the superiority of our proposed model and visualized the learned attention map to make the model interpretable. As future work, we will apply \name to other spatial-temporal prediction tasks, such as wind power forecasting~\cite{jiang2018buaa_bigscity}. In addition, we will explore the pre-training techniques in traffic prediction to solve the problem of insufficient data.
\clearpage

\clearpage
\section*{Acknowledgments}
This work was supported by the National Key R\&D Program of China (Grant No. 2019YFB2102100). Prof. Wang’s work was supported by the National Natural Science Foundation of China (No. 72171013, 82161148011, 72222022), the Fundamental Research Funds for the Central Universities (YWF-22-L-838) and the DiDi Gaia Collaborative Research Funds. Prof. Zhao’s work was supported by the National Natural Science Foundation of China (No. 62222215).

\bibliography{aaai23.bib}

\end{document}